\documentclass[sigconf]{acmart}
\AtBeginDocument{%
  }

\copyrightyear{2026}
\acmYear{2026}
\setcopyright{cc}
\setcctype{by}

\acmDOI{XXXXXXX.XXXXXXX}
\acmISBN{979-8-4007-2539-5/2026/11}
\acmConference[EARL Workshop@CIKM'26]{The 3rd Workshop on Evaluating and Applying Recommender Systems with Large Language Models at Conference on Information and Knowledge Management}{November 7--11, 2026}{Rome, Italy}

\usepackage{lipsum}
\usepackage{graphicx}
\usepackage{multirow}
\usepackage{tabularx}
\usepackage{booktabs}
\usepackage{makecell}
\usepackage{array}
\usepackage{float}
\usepackage{placeins}
\usepackage{bm}
\usepackage{amsmath}

\newcommand{\ours}{SearCl}
\begin{document}

\title{Dialogue Generation for Conversational Recommendation Using User Search Interactions and Knowledge Graphs
}

\author{Minsung Hwang}
\affiliation{%
  \institution{School of Computing, KAIST}
  \city{Daejeon}
  \country{Republic of Korea}
}
\email{hminsung@kaist.ac.kr}

\author{Donggyu Yoon}
\affiliation{%
  \institution{School of Computing, KAIST}
  \city{Daejeon}
  \country{Republic of Korea}
}
\email{justinyoonwk@kaist.ac.kr}

\author{Gowoon Hwang}
\affiliation{%
  \institution{LG Electronics}
  \city{Seoul}
  \country{Republic of Korea}
}
\email{gowoon.hwang@lge.com}

\author{Kwanghee Jang}
\affiliation{%
  \institution{LG Electronics}
  \city{Seoul}
  \country{Republic of Korea}
}
\email{kwanghee.jang@lge.com}

\author{Joyce Jiyoung Whang}
\authornote{Corresponding author}
\affiliation{%
  \institution{Department of AI Computing, KAIST}
  \city{Daejeon}
  \country{Republic of Korea}
}
\email{jjwhang@kaist.ac.kr}

\begin{abstract} 
Large-scale conversational recommendation datasets typically require manual dialogue construction or annotation. We propose \ours\ which automatically generates conversational recommendation dialogues grounded in user \underline{sear}ch and \underline{cl}ick behaviors. \ours\ segments user interactions into query sessions and derives each recommendation target from content selected after a search query. A configurable probabilistic state machine controls the dialogue flow, while two LLM agents acting as the seeker and recommender generate utterances according to turn-level constraints. To support recommendation, the recommender is provided with knowledge graphs constructed from metadata about the content items used in each dialogue, capturing their attributes and relationships. Using \ours, we generate 3,538 dialogues from content search and click interaction data, covering 1,731 unique target items. Under an LLM-as-a-judge protocol, the resulting dialogues achieve higher quality scores than dialogues from existing benchmarks. These results demonstrate that \ours\ enables controllable and knowledge-grounded dataset construction without human annotations. 
\end{abstract}




\maketitle

\section{Introduction}
\label{sec:introduction}

Conversational recommender systems (CRSs) support users in discovering items through multiple conversational turns that elicit preferences, refine requirements, and explain recommendations~\cite{memocrs,step,harpo,taira,cope}. Developing such systems requires dialogue datasets containing realistic user intents, appropriate recommendation targets, and coherent conversational flows~\cite{redial,inspired,durecdial,opendialkg}. Existing CRS datasets, including ReDIAL~\cite{redial} and INSPIRED~\cite{inspired}, have played an important role in advancing the field of CRS, but human dialogue collection and annotation are costly and difficult to scale. LLM-based generation can reduce this cost and improve scalability, but existing approaches often rely on manually designed dialogue templates and preference profiles derived from ratings or reviews~\cite{llmredial}. This dependence can limit flexibility and make it difficult to account for users' search-and-click behavior.

\begin{figure}[!t]
    \centering
    \includegraphics[width=\columnwidth]{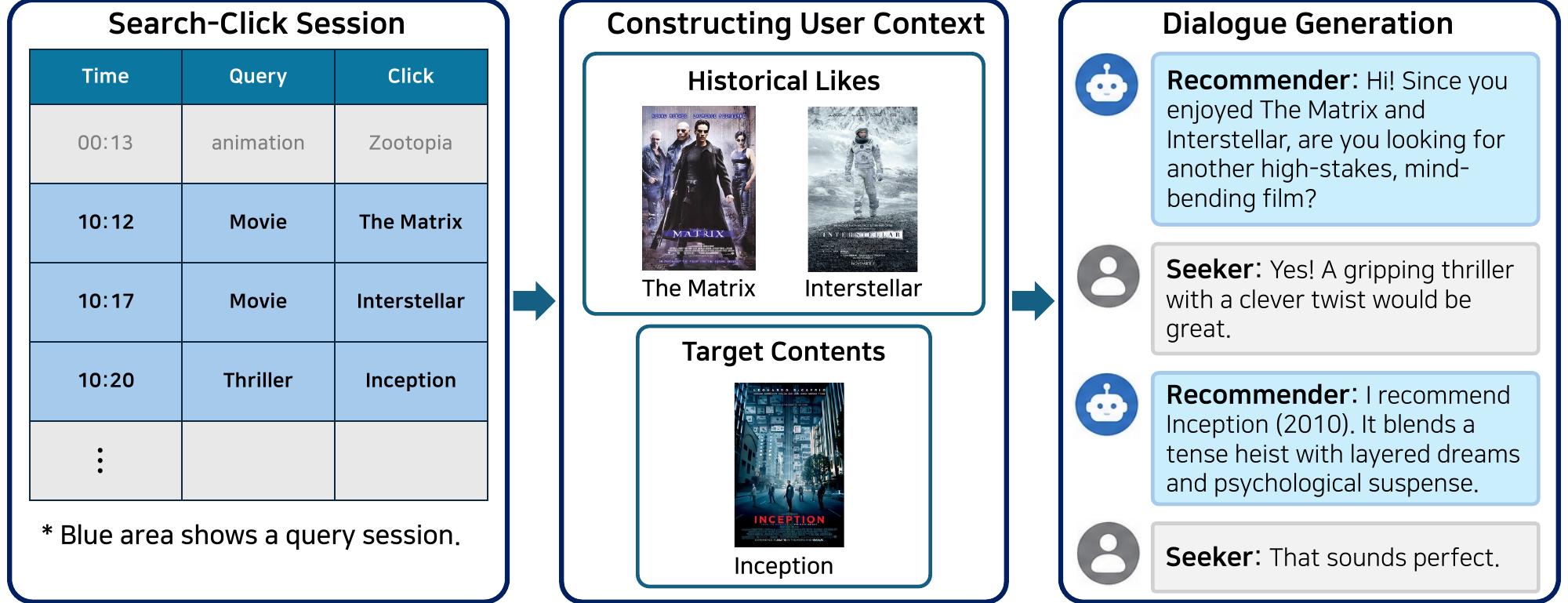}
    \caption{An example of generating a recommendation dialogue from search-click interactions. The interactions are converted into a user profile and a recommendation target, which are then used to construct the dialogue.}
    \Description{}
    \label{fig:pipeline_overview}
\end{figure}

We propose \ours, a pipeline for automatically constructing dialogues for conversational recommendation from user \underline{sear}ch and \underline{cl}ick behavior. As illustrated in Figure~\ref{fig:pipeline_overview}, the pipeline groups content search and click interactions into query sessions, which represent continuous periods of user interaction, and transforms each session into a conversational recommendation scenario. Within each query session, content items clicked before a target query are used to construct the user context, while the item clicked immediately after that query is designated as the recommendation target. These inputs are then converted into a multi-turn dialogue between seeker and recommender LLM agents. A configurable probabilistic state machine determines the conversational structure, and each agent generates utterances according to constraints specified for each turn. To support recommendation explanations, we use a content knowledge graph (KG) as grounding information. KGs provide a flexible representation that can accommodate emerging entities, relations, and facts~\cite{ingram,krepe,maypl}. They can also integrate multiple modalities, including numeric attributes and visual--textual descriptions~\cite{vista,hynt}, providing richer evidence for recommendation explanations. Accordingly, the recommender receives bounded two-hop subgraphs extracted from the KG as grounding context. \ours\ also applies automatic filtering to remove target title leakage and trivial cases where the query matches the target title. Using \ours, we generate dialogues from real-world interaction data, demonstrating that the proposed method can produce dialogues grounded in user behavior with controlled conversational structures and without manual construction. 
Our main contributions are as follows:
\begin{itemize}
    \item We propose \ours, a method for generating conversational recommendation dialogues from observed content search and click interaction data.
    \item We support flexible dialogue design by planning dialogue turns through a configurable probabilistic state machine, whose transition probabilities and available turn types can be adjusted to produce different conversational structures.
    \item We incorporate content knowledge graphs into dialogue generation, enabling recommendation explanations grounded in structured and auditable evidence.
\end{itemize}

\section{Related Work}
\label{sec:related_work}

A variety of datasets have been developed to support research on conversational recommender systems. ReDIAL contains more than 10,000 human conversations centered on movie recommendations~\cite{redial}, while INSPIRED contains 1,001 human conversations annotated with sociable recommendation strategies such as expressing personal opinions and providing encouragement~\cite{inspired}. Although these datasets provide valuable examples of human recommendation dialogue, their dependence on human participation makes expansion to larger scales, additional domains, and new scenarios costly. To address this limitation, our work focuses on an automated LLM-based approach that constructs behavior-grounded recommendation dialogues without human annotations.

LLM-based dialogue generation has emerged as a scalable alternative to human data collection~\cite{llmredial,muse,distillrecdial}. LLM-REDIAL fills manually designed dialogue templates with users' historical interactions and reviews~\cite{llmredial}. MUSE employs a multi-agent multimodal LLM framework to synthesize dialogues from scenario-grounded user profiles~\cite{muse}. DistillRecDial uses knowledge distillation to generate dialogues with smaller LLMs while simulating diverse user types~\cite{distillrecdial}. These methods enable large-scale dialogue generation, but they neither incorporate observed search and click behavior into recommendation scenarios nor provide recommendation explanations grounded in KG evidence. In contrast, we propose a dialogue generation method which constructs recommendation scenarios from observed search-click interactions and provides recommendation explanations based on KG constructed by metadata.
\section{Methodology: \ours}
\label{sec:method}

\begin{figure}[t]
    \centering
    \includegraphics[width=\columnwidth]{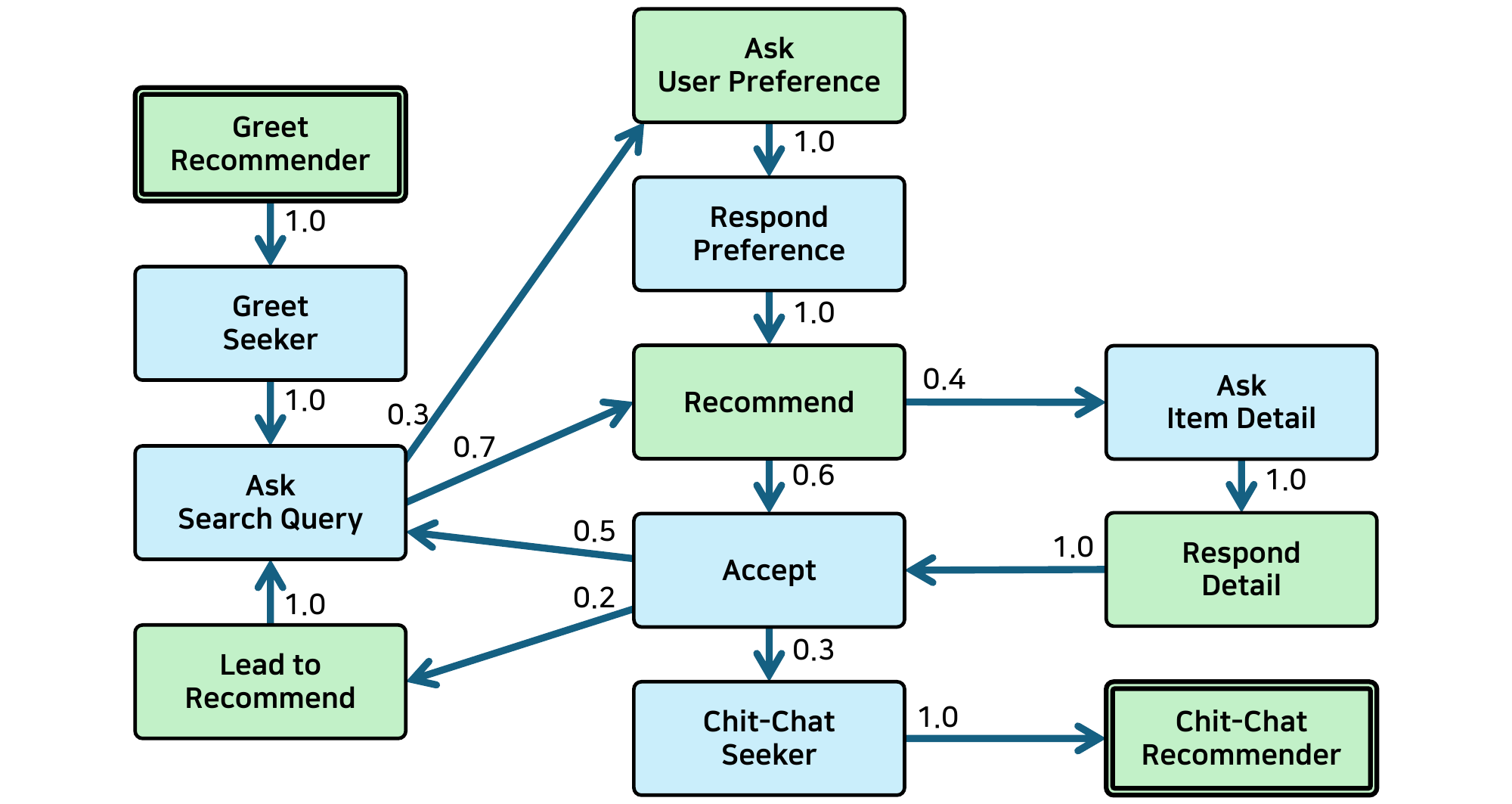}
    \caption{The probabilistic state machine configuration used to plan all dialogues in this study. Blue nodes denote seeker subgoals, whereas green nodes denote recommender subgoals. \textsc{Greet Recommender} and \textsc{Chit-Chat Recommender} are the initial and terminal states, respectively. Edge labels show the corresponding transition probabilities.}
    \label{fig:state_machine}
\end{figure}

We propose \ours, a method for constructing conversational recommendation dialogues from content search and click interactions. \ours\ first groups these interactions into query sessions, where each observed association between a search query and a clicked item defines a recommendation scenario grounded in the user's search behavior and selected content. For each scenario, \ours\ constructs the user context from content items clicked before the corresponding target query within the same query session, uses a configurable probabilistic state machine to plan each dialogue turn, and employs LLM agents assigned the roles of seeker and recommender to generate utterances according to the turn plan. To provide explanations grounded in content metadata, \ours\ constructs a knowledge graph representing relations between content items and their attributes and provides it to the recommender agent. Finally, \ours\ filters out dialogues containing target leakage or trivial query--title matches.

\subsection{Query Session Construction}
\label{sec:scenario}

We assume that the source data contain anonymized user identifiers, timestamps, search queries, content identifiers, domains, and clicked items. We retain movie and TV-show interactions that can be linked to metadata such as genres and cast members. Interactions are ordered by user and grouped into query sessions based on positive inter-event gaps ($\Delta t$). Following prior work on web sessionization~\cite{halfaker2015user}, we fit a two-component Gaussian mixture to ($x=\log(1+\Delta t)$). We assume that the component with the smaller mean in log-gap space captures pauses within a session, whereas the other component captures longer periods of inactivity between sessions. The inactivity threshold is defined as the first crossover of the weighted component densities between their means, converted back to seconds and clipped to 5--120 minutes. A user-specific model requires at least five gaps, corresponding to six interactions; otherwise, we use the conventional 30-minute threshold. Under this rule, 5,632 of 65,695 users (8.6\%) received user-specific thresholds, while the remaining 91.4\% used the fallback. A new session begins when a consecutive-event gap exceeds the applicable threshold.

After constructing the query sessions, we distinguish historical likes from recommendation targets based on the temporal order of interactions within each session. Content items clicked before a target query are used to construct the historical preference context, while the item clicked immediately after the query is designated as the recommendation target. The query determines how the seeker expresses the current recommendation need, and the corresponding target is introduced by the recommender at the designated recommendation turn. Queries without an associated click are retained only as auxiliary context. This construction grounds each scenario in an observed transition from an information need to a selected item rather than in a manually assigned preference.

\begin{table*}[t]
\centering
\small
\caption{An example dialogue generated by \ours\ from a query session containing the search query \textit{movie}. The user context includes \textit{Ttoltoli: The Secret of the Baby Dinosaur} as a previously liked item, and \textit{The Super Mario Bros. Movie (2023)} is the recommendation target. The original Korean utterances are translated into English for presentation.}
\label{tab:turnplan-variant-b}
\begin{tabular}{@{}p{0.19\linewidth}p{0.74\linewidth}@{}}
\toprule
\textbf{State(s)} & \textbf{Utterance} \\
\midrule
\texttt{GREET\_RECOMMENDER} & Hello! You previously enjoyed children's animation such as \textit{Ttoltoli: The Secret of the Baby Dinosaur}. Would you like to watch something similar today, or should I suggest something with a different atmosphere? \\
\shortstack[l]{\texttt{GREET\_SEEKER}\\$\rightarrow$ \texttt{ASK\_SEARCH\_QUERY}} & Hello! Yes, please choose something for me again today. I am looking for a light and entertaining movie that is not too serious and is suitable for watching with my family. Could you recommend one? \\
\texttt{ASK\_USER\_PREFERENCE} & For something that you can comfortably watch with your family, animation or comedy would be a good fit. Would you prefer something completely bright and cheerful, or a story with a little more adventure? \\
\texttt{RESPOND\_PREFERENCE} & I would prefer something bright and cheerful! \\
\texttt{RECOMMEND} & In that case, I recommend \textit{The Super Mario Bros. Movie (2023)}. Its combination of comedy, animation, and adventure makes it bright and entertaining. With voice performances by Jack Black and Anya Taylor-Joy, it is also a fun choice for the whole family. \\
\shortstack[l]{\texttt{ACCEPT}\\$\rightarrow$ \texttt{CHITCHAT\_SEEKER}} & Great, I will watch that one! Its bright and cheerful atmosphere sounds perfect for a relaxing movie night with my family. If we enjoy it, I will ask for something similar next time. \\
\texttt{CHITCHAT\_RECOMMENDER} & I hope you have a great time watching it with your family! If you enjoy it, I can recommend another bright and lighthearted movie next time. \\
\bottomrule
\end{tabular}
\end{table*}

\subsection{Dialogue Turn Planning}
\label{sec:template_construction}

For each query session, we plan the dialogue at the turn level by sampling a sequence of subgoals using a predefined probabilistic state machine. Each node represents a seeker or recommender subgoal, and each edge represents a possible transition. Consecutive subgoals assigned to the same speaker are merged into a single dialogue turn. For each resulting turn, the plan specifies the active speaker, one or more dialogue subgoals, the content items to be used for utterance generation, and turn-specific generation constraints. The available subgoals include greeting, expressing a search need, eliciting preferences, requesting item details, recommending an item, accepting a recommendation, and closing the conversation. Both the transition probabilities and the available subgoals can be configured according to the properties of the source dataset and the objectives of dialogue generation, enabling flexible design.

Figure~\ref{fig:state_machine} presents the values we used as one example configuration. The main flow begins with a greeting and a search request, followed by either a direct recommendation or an optional preference-elicitation exchange. After receiving a recommendation, the seeker may accept it or request additional details. An accepted recommendation may lead to another query or a closing exchange, allowing multiple query--target pairs from the same query session to appear in one dialogue. Once all targets have been covered, the state machine prevents further transitions to additional recommendations. The resulting turn plan guides the recommendation flow while allowing the LLM agents flexibility in utterance generation.

\subsection{Automatic Dialogue Generation}
\label{sec:generation}

Following the turn plan, separate LLM agents assigned the roles of seeker and recommender generate the dialogue one turn at a time. At each turn, the active agent receives its role, the dialogue history, the current subgoal, the content items to be used in generating the utterance, and turn-specific constraints. The agent then generates a single utterance. Search-request turns prohibit disclosure of the target title, whereas recommendation turns require the designated target and relevant supporting information. This turn-by-turn process enables the agents to produce natural and contextually coherent utterances while closely following the plan.

The recommender receives the properties of the content items designated for each turn, including plot summaries and other metadata. The recommender additionally receives subgraphs extracted from a content KG built from structured content metadata. The KG uses relations to connect content items with metadata entities such as genres, people, series, and semantic tags, as well as with related content items that share similar textual or visual features. It contains 792,395 entities, 5,618,485 triplets, and 14 relations. For every content item appearing in the user context, as a recommendation target, or in the turn plan, we extract a bounded two-hop subgraph. To prevent the input KG from expanding excessively, we limit the number of neighbors retained at each hop. The union of these subgraphs forms the dialogue-level KG context used by the recommender throughout the dialogue. This context supplies structured evidence for recommendation explanations while limiting prompt length. The KG defines an auditable evidence space but does not guarantee the factuality of every generated statement; factual support is therefore evaluated separately.

Finally, we apply automatic filters to remove dialogues that violate the intended conversation flow or make the recommendation task trivial. Specifically, we discard dialogues in which the target title is revealed before the designated recommendation turn. We also remove query--target pairs for which the original search query directly matches the target title, as these cases reflect direct item lookup rather than recommendation. These post-generation filters complement the turn-specific constraints used during generation: the constraints discourage target leakage, while the filters remove any remaining violations. Table~\ref{tab:turnplan-variant-b} presents a retained dialogue generated from content search and click interactions. The search query is transformed into a conversational request, while the clicked item is introduced only at the designated recommendation turn.

\begin{figure}[t]
    \centering
    \includegraphics[width=\columnwidth]{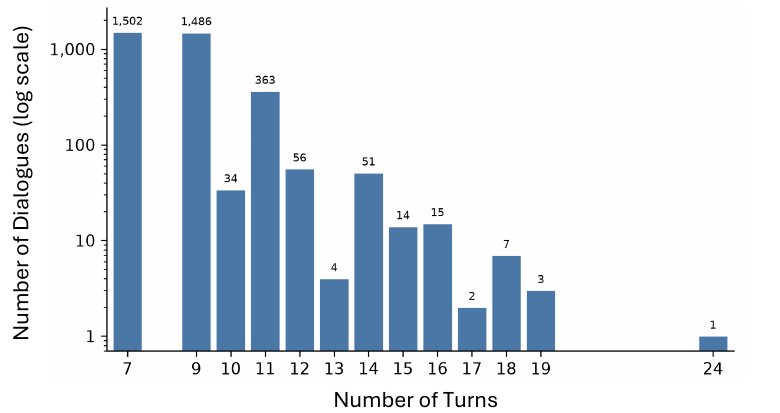}
    \caption{Distribution of dialogue counts by turn-plan length.}
    \label{fig:plan_length_histogram}
\end{figure}

\section{Experiments}
\label{sec:experiments}

We evaluate \ours\ using TV usage records that were anonymized prior to being provided to the researchers by a leading consumer electronics company in South Korea. The records contain search and viewing activities related to movie and TV show content. Our framework can be applied to other datasets containing search and viewing interactions that can be linked to metadata. Using these records, we generated dialogues using \ours\ with \texttt{gpt-5.4-mini}~\cite{gpt54mini}.

\subsection{Dataset Statistics}
\label{sec:dataset_statistics}

\begin{figure}[t]
    \centering
    \includegraphics[width=\columnwidth]{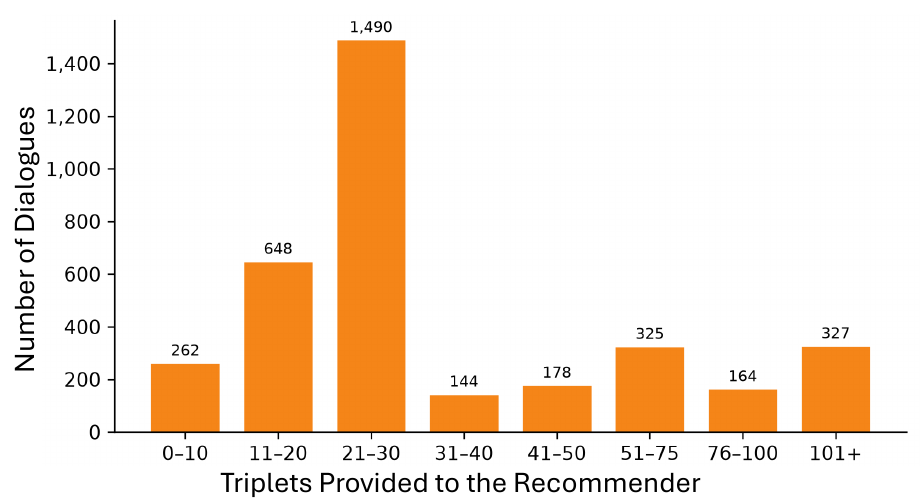}
    \caption{Distribution of the number of triplets provided to the recommender for each dialogue.}
    \label{fig:triplet_count_histogram}
\end{figure}

Using \ours, we initially generated 10,000 candidate dialogues from content search and click interaction data. After filtering out dialogues containing target leakage or trivial query--title matches, 3,538 dialogues were retained, covering 1,731 unique recommendation targets. The retained dialogues contain an average of 8.58 turns and follow 164 unique turn plans. Figure~\ref{fig:plan_length_histogram} presents their distribution by turn-plan length, while Figure~\ref{fig:triplet_count_histogram} presents their distribution across binned numbers of knowledge graph triplets used for dialogue generation. Although the resulting dataset was constructed to evaluate \ours, it can also serve as a resource for fine-tuning and evaluating conversational recommender system models.

\subsection{Dialogue Quality Evaluation}
\label{sec:quality_evaluation}

We compare our dataset with 100-dialogue samples from ReDIAL~\cite{redial} and INSPIRED~\cite{inspired}. All corpora are evaluated using the same LLM-as-a-judge protocol~\cite{llmjudge} with \texttt{gpt-5.6-luna}~\cite{gpt56}. Following LLM-REDIAL~\cite{llmredial}, we evaluate utterances in terms of \textit{Fluency}, \textit{Informativeness}, \textit{Logicality}, and \textit{Coherence} using a 0--2 scale. We additionally introduce \textit{Constraint adherence} and \textit{Grounding}, also scored from 0 to 2, to evaluate properties of our method. For conversation-level evaluation, we introduce \textit{Completeness}, \textit{History utilization}, and \textit{Overall quality}, which are scored on a standard five-point Likert scale.

At the utterance level, \textit{Fluency} measures grammatical correctness and naturalness, while \textit{Informativeness} assesses whether an utterance provides useful and specific information. \textit{Logicality} evaluates internal consistency, and \textit{Coherence} measures how well the utterance follows the preceding dialogue context. \textit{Constraint adherence} determines whether the utterance satisfies the conditions of the turn plan, including the designated speaker and subgoal, whether the target content item is included, and whether the utterance does not include test leakage. \textit{Grounding} evaluates whether recommendations are supported by the user context, content metadata, or KG evidence. At the conversation level, \textit{Completeness} measures whether the dialogue contains the expected recommendation stages in an appropriate order. \textit{History utilization} assesses how effectively the interaction history is incorporated, while \textit{Overall quality} captures the dialogue's overall naturalness and consistency as a dialogue.

As shown in Table~\ref{tab:quality_comparison}, our dataset receives higher average scores than the sampled ReDIAL~\cite{redial} and INSPIRED~\cite{inspired} dialogues across all comparable dimensions. At the utterance level, the generated dialogues achieve high fluency, logicality, coherence, and constraint adherence, while also outperforming both comparison corpora in informativeness and grounding. At the conversation level, the largest improvement is observed in overall quality. These results indicate that the proposed method produces coherent dialogues that follow the intended recommendation structure and effectively use the grounding information provided to the generator.

\begin{table}[t]
\caption{LLM-based quality evaluation results of dialogues from \ours\ and sampled dialogues from ReDIAL~\cite{redial} and INSPIRED~\cite{inspired}. Utterance-level scores are averaged over all individual utterances, whereas conversation-level scores are averaged over all conversations.}\label{tab:quality_comparison}
\centering
\small
\setlength{\tabcolsep}{2.2pt}
\begin{tabular}{@{}llccc@{}}
\toprule
\textbf{Level}
    & \textbf{Metric}
    & \textbf{\ours}
    & \textbf{ReDIAL}
    & \textbf{INSPIRED} \\
\midrule
\multirow{6}{*}{\shortstack[l]{Utterance\\(0--2)}}
    & Fluency              & \textbf{\underline{1.98}} & 1.67 & 1.71 \\
    & Informativeness      & \textbf{\underline{1.35}} & 0.99 & 1.03 \\
    & Logicality           & \textbf{\underline{1.95}} & 1.86 & 1.86 \\
    & Coherence            & \textbf{\underline{1.96}} & 1.89 & 1.92 \\
    & Constraint adherence & \textbf{\underline{1.99}} & --   & --   \\
    & Grounding            & \textbf{\underline{1.81}} & 1.52 & 1.33 \\
\midrule
\multirow{3}{*}{\shortstack[l]{Conversation\\(1--5)}}
    & Completeness          & \textbf{\underline{4.67}} & 3.35 & 3.99 \\
    & History utilization   & \textbf{\underline{4.54}} & 4.03 & 3.48 \\
    & Overall quality       & \textbf{\underline{4.62}} & 3.18 & 3.53 \\
\bottomrule
\end{tabular}
\end{table}

\subsection{Ablation Study}
\label{sec:kg_ablation}

To isolate the effect of KG context, we generate a pair of dialogues for every query session in the dataset. Each pair uses the same user context, recommendation target, turn plan, generation model, and interaction context, with KG context as the only experimental variable. In the condition without the KG, the recommender still receives all properties of the recommended content, including its plot summary and other metadata; only the relational context derived from the KG is removed. We compare the overall quality scores between the two conditions using a two-sided paired $t$ test. Providing KG context produces a statistically significant increase in mean overall quality, from 4.556 without KG context to 4.620 with it ($p=6.3 \times 10^{-8}$). This result indicates that relational information about content entities helps the recommender produce more appropriate and well-supported responses beyond what can be achieved using content properties alone.

\section{Conclusion}
\label{sec:conclusion}

We present \ours, which constructs conversational recommendation datasets from search and click interactions. \ours\ grounds recommendation targets in observed user actions, plans dialogue turns through a configurable probabilistic state machine, and uses a content knowledge graph to support recommendation explanations. Our experiments demonstrate that \ours\ generates diverse and coherent dialogues at scale without manual dialogue writing, while achieving higher average quality scores than representative conversational recommendation benchmarks. The transition probabilities and the set of turn types considered during planning are configurable, allowing \ours\ to accommodate different dialogue patterns and interaction settings. 

The flexibility of \ours\ allows it to incorporate richer behavioral signals that distinguish sustained preferences from temporary interest or curiosity and provide evidence of negative preferences. These signals can support dialogue plans involving rejection, dissatisfaction, and unsuccessful recommendations. This configurability also provides a natural extension toward replacing globally specified transition probabilities with context-dependent ones. Recent work has used the current query and dialogue history to determine whether a system should continue the interaction or provide a recommendation~\cite{memn2n}. Applying this idea to the finer-grained subgoals in \ours\ could produce dialogue plans that dynamically adapt to the current query, dialogue state, and available user context.

\section*{GenAI Usage Disclosure}

In this work, we used generative AI tools to generate synthetic conversational recommendation data and conduct LLM-based quality evaluation. As described in Sections~\ref{sec:generation} and~\ref{sec:quality_evaluation}, OpenAI's gpt-5.4-mini was used to generate utterances for the seeker and recommender agents under author-defined roles, dialogue histories, turn-level subgoals, content constraints, and a knowledge graph. OpenAI's gpt-5.6-luna was used as a judge to evaluate \ours\ and the comparison datasets using author-defined evaluation criteria and prompts. The underlying search-and-click logs, query-session segmentation, user contexts, recommendation targets, dialogue plans, and KG subgraphs were obtained or constructed programmatically. The generated dialogues were processed using predefined constraints and post-generation filters.

Additionally, generative AI tools were used solely for proofreading, grammar correction, and minor wording and stylistic edits to improve the readability of the manuscript. They were not used to develop the research ideas, methods, or scientific claims. The authors reviewed all AI-assisted manuscript changes and take responsibility for the final content of this work, including its text, claims, generated data, evaluation procedure, and conclusions.

We did not use generative AI tools to formulate or refine research hypotheses, design or provide feedback on the research methodology or experiments, implement the proposed methods, preprocess or clean the source interaction logs, construct query sessions, dialogue plans, or knowledge graphs, perform statistical tests, or interpret the experimental results.

\section*{Acknowledgments}
This work was supported by the National Research Foundation of Korea (NRF) grant funded by the Korean government (MSIT) (RS-2025-00559066).

\bibliographystyle{ACM-Reference-Format}
\bibliography{cikm_2026_workshop}


\end{document}